\documentclass[letterpaper, 10 pt, conference]{ieeeconf}  

\IEEEoverridecommandlockouts                              

\usepackage{graphics} 
\usepackage{epsfig} 
\usepackage{mathptmx} 
\usepackage{times} 
\usepackage{amsmath} 
\usepackage{amssymb}  

\usepackage{graphicx}
\usepackage{subcaption}

\newcommand{\Sup}{$\checkmark$}  
\newcommand{\Par}{$\triangle$}   
\newcommand{\No}{$\times$}       

\usepackage{booktabs}

\title{\LARGE \bf
Chrono-Gymnasium: An Open-Source, Gymnasium-Compatible Distributed Simulation Framework
}

\author{Bocheng Zou$^{1}$, Harry Zhang$^{1}$, Khailanii Slaton$^{1}$, Jingquan Wang$^{1}$, Derrick Ruan$^{1}$, \\ Huzaifa Mustafa Unjhawala$^{1}$, Radu Serban$^{1}$ and Dan Negrut$^{1}$
	\thanks{$^{1}$University of Wisconsin-Madison}
}

\graphicspath{{./RAL_ChronoGym/}{../../../../local-image-archive/journals/2026/RAL_ChronoGym/}}

\begin{document}

	\maketitle
	\thispagestyle{empty}
	\pagestyle{empty}

	\begin{abstract}
		
		High-fidelity physics simulation is essential for closing the sim-to-real gap in robotics and complex mechanical systems. However, the computational overhead of high-fidelity engines often limits their use in data-intensive tasks like Reinforcement Learning (RL) and global optimization. We introduce Chrono-Gymnasium, a distributed computing framework that scales the high-fidelity multi-body dynamics of Project Chrono across large-scale computing clusters. Built upon the Ray framework, Chrono-Gymnasium provides a standardized Gymnasium interface, enabling seamless integration with modern machine learning libraries while providing built-in synchronization and messaging primitives for distributed execution. We demonstrate the framework's capabilities through two distinct case studies: (1) the training of an RL agent for autonomous robotic navigation in complex terrains, and (2) the Bayesian Optimization of a planetary lander's design parameters to ensure landing stability. Our results show that Chrono-Gymnasium reduces wall-clock time for high-fidelity simulations without sacrificing physical accuracy, offering a scalable path for the design and control of complex robotic systems.

	\end{abstract}
	
	\section{INTRODUCTION}

The demand for high-fidelity simulation in robotics has seen a paradigm shift, driven largely by the success of data-driven approaches such as Reinforcement Learning (RL) and complex engineering design optimization. While simulators allow for safe, repeatable, and accelerated training, the ``reality gap" remains a significant hurdle. Closing this gap often requires high-fidelity physics engines capable of simulating complex rigid and flexible multi-body dynamics, non-smooth contacts, terramechanics, fluid-structure interaction phenomena, actuator dynamics and control. However, the increased physical accuracy comes at a computational cost, which often renders single-node simulations insufficiently expeditious for large-scale learning or optimization tasks.

Project Chrono~\cite{projectChronoWebSite} has emerged as a versatile, open-source physics engine for these high-fidelity requirements, particularly in off-road mobility and complex multi-physics robotics applications. Yet, transitioning from a standalone Chrono simulation to a massively parallelized environment remains a non-trivial engineering challenge. Researchers are often forced to write custom, ad-hoc wrappers to handle inter-process communication, data synchronization, and resource management across distributed clusters.

To address these challenges, we present Chrono-Gymnasium, a distributed computing framework designed to scale high-fidelity physics simulations across heterogeneous clusters. Chrono-Gymnasium builds a bridge between the rigorous multi-physics simulation of Project Chrono and the scalable distributed execution capabilities of the Ray framework \cite{moritz2018ray}. By providing a standardized Gymnasium (formerly OpenAI Gym) interface \cite{towers2024gymnasium}, Chrono-Gymnasium enables researchers to integrate high-fidelity simulations into existing machine learning pipelines.

The remainder of this paper is organized as follows: Section II reviews related work in distributed simulation; Section III details the architecture of Chrono-Gymnasium; Section IV describes the experimental setup and results for our case studies; and Section V concludes with future directions.

	\section{RELATED WORK}

\begin{table*}[t]
	\centering
	\vspace{0.5em}
	\caption{Comparison of modern physics simulators across multi-physics modeling capabilities (Rigid, FEA, and SPH) and parallelization support (Single-experiment vs. Vectorized Cross-experiment). The table illustrates how Chrono-Gymnasium bridges the gap between high-fidelity engines like Project Chrono and the scalable, high-throughput execution required for reinforcement learning and design optimization. Legend: \Sup\ supported, \Par\ partial, \No\ not supported.}
	\label{tab:related_sim_comp}
	\begin{tabular}{lccccc}
		\toprule
		& \multicolumn{3}{c}{Capabilities} & \multicolumn{2}{c}{Parallelization} \\
		\cmidrule(lr){2-4}\cmidrule(lr){5-6}
		Simulator & Rigid & FEA & SPH & Single-experiment & Cross-experiment (Vectorized Env) \\
		\midrule
		Gazebo \cite{koenig2004design}    & \Sup & \No & \No  & \Par & \No \\
		CoppeliaSim \cite{coppeliaSim}  & \Sup & \No & \No  & \Par & \No \\
		Bullet \cite{coumans2015bullet}   & \Sup & \Sup & \No  & \Par & \No \\
		SPHinXsys \cite{zhang2021sphinxsys} & \Sup & \No & \Sup & \Sup & \No \\
		Drake  \cite{drake}  & \Sup & \Par & \No& \Par & \Par\footnotemark[1] \\
		Isaac Lab \cite{mittal2025isaac}     & \Sup & \Sup  & \No & \Sup & \Sup \\
		Brax \cite{freeman2021brax}    & \Sup & \No & \No  & \Sup & \Sup \\
		Genesis \cite{Genesis}                & \Sup & \Sup & \Sup & \Sup & \Sup \\
		\midrule
		Project Chrono  \cite{chronoOverview2016} & \Sup & \Sup & \Sup &  \Par \footnotemark[2]& \No \\
		Project Chrono  with Chrono-Gymnasium & \Sup & \Sup & \Sup & \Par \footnotemark[2]& \Sup \\
		\bottomrule
	\end{tabular}
	
\end{table*}

\subsection{Simulators with Native Parallel Support}

Classic general-purpose simulators such as Gazebo \cite{koenig2004design}, CoppeliaSim \cite{coppeliaSim}, and Bullet \cite{coumans2015bullet} emphasize broad model compatibility and fast iteration for rigid-body robotics. However, as indicated in Table \ref{tab:related_sim_comp}, these systems typically provide only limited intra-experiment parallelism (e.g., partial multithreading) and do not natively support cross-experiment vectorized execution for large-scale experience collection. As a result, they cannot make good use of modern multi-core CPUs and large-scale computing clusters.

Meanwhile, multi-physics libraries such as SPHinXsys \cite{zhang2021sphinxsys}, although they provide richer physics, such as native support for SPH \cite{unjhawala2025physics}, with good single-experiment parallelization support, they are still not typically packaged as turnkey, vectorized RL simulators for embodied AI applications.

Recent simulators targeting robot learning treat large-scale parallel rollout as a first-class design objective. Isaac Lab \cite{mittal2025isaac} and Brax \cite{freeman2021brax} exemplify this trend: Table \ref{tab:related_sim_comp} indicates that both support efficient single-experiment stepping and, critically, cross-experiment vectorized execution, enabling high-throughput data generation by running many environments in parallel. This tight coupling between physics stepping and batched environment execution can dramatically reduce wall-clock training time, especially when the learning stack is organized around parallel rollouts. However, they both focus mostly on rigid body dynamics, with limited or no support for fluid and soft body first principles simulation solutions.

Genesis \cite{Genesis} is a recent simulator targeting robot learning and embodied AI with native support for rigid-body and SPH physics and for both single-experiment and cross-experiment (vectorized) parallel execution. It thus bridges physics breadth and scalable vectorized RL, offering a turnkey environment for high-throughput learning with richer physics than rigid-body--only engines. However, Chrono is built to interface with engineering and software workflows. Chrono can employ, via the Funcitonal Mockup Interface (FMI) \cite{raduFMIChrono2025}, external third party libraries to produce a system-level simulation that can include hardware and/or human in the loop. 

Chrono-Gymnasium complements the afore-mentioned simulators by enabling scalable learning and optimization while retaining the modeling depth of Project Chrono, whose DNA is rooted in high-fidelity multi-physics simulation for real-world engineering applications, see, for instance, \cite{gravOffset2025,schepelmann2025overview,Rezich-VPER-Chrono-2025}. Looking at the solutions listed in Table \ref{tab:related_sim_comp}, Chrono-Gymnasium and Drake trade speed for accuracy, while the remaining ones trade accuracy for speed. Arguably, these two camps are complementary, and they are both well-suited for different use cases. 

\footnotetext[1]{Limited supported for MonteCarloSimulation and RL}
\footnotetext[2]{Only in supported systems, such as SynChrono, DEM, FSI, ChSystemMulticore}

\subsection{Parallelizing Project Chrono}
Prior efforts to parallelize Chrono-based workflows largely fall into two categories. The first category accelerates a \emph{single} simulation instance through parallel algorithms or distributed execution. Project Chrono provides parallel computing support through several built-in modules for a limited set of physical phenomena.  \texttt{Chrono::MultiCore} module provides parallel support for rigid body simulation, while \texttt{SynChrono} employs MPI-based coordination to enable scalable simulation of multiple interacting agents (vehicles and/or robots) within one coupled scenario \cite{taves2020synchrono}.

The second category scales \emph{across} simulation instances. Existing work utilizes general HPC tools like Slurm to achieve embarrassingly parallel execution of many independent runs for Monte Carlo analysis, surrogate modeling, parameter sweeps, and black-box optimization \cite{zhang2026data}. However, this setting is only applicable to downstream applications that don't require synchronization and communication between experiment instances.  Gym-Chrono \cite{benatti2021pychrono} , a derivative of Project Chrono, offers the capability to run multiple experiments to support reinforcement learning, however, the experiments run serially and are therefore limited by single-thread CPU performance.

Chrono-Gymnasium builds on these prior directions by focusing explicitly on the \emph{workflow layer} needed for autonomy development: it standardizes Chrono simulations behind the Gymnasium interface while providing distributed execution primitives that make scaling to many parallel rollouts routine. This shifts parallelization effort from ad-hoc experiment code into a reusable framework, reducing the marginal cost of running RL, Bayesian optimization, or large design studies on high-fidelity Chrono models.

\subsection{Distributed Computing Frameworks for Simulation and Learning}
Distributed simulation and learning have traditionally relied on message-passing interfaces (MPI) \cite{graham2005open} and HPC schedulers, which offer fine control and high performance for tightly coupled workloads. MPI-style solutions are well-suited for synchronized multi-agent simulation, but they often require explicit management of communication, fault tolerance, and resource allocation, and they provide limited out-of-the-box integration with Python-centric ML ecosystems.

In contrast, modern distributed computing frameworks such as Ray \cite{moritz2018ray} provide higher-level abstractions (tasks, actors, object stores) that simplify scaling Python applications across heterogeneous clusters. Ray has also become a common substrate for distributed RL systems and experience collection due to its composable parallelism and flexible resource management. Chrono-Gymnasium leverages these advantages by embedding Chrono simulation instances within Ray-managed workers while exposing a Gymnasium-compliant API to downstream learning and optimization libraries.

A key design goal is to make distributed execution \emph{transparent} to the autonomy developer: the same environment code should run on a laptop (with local parallelism) or on a cluster (with many distributed workers), while maintaining consistent semantics for stepping, resets, and data collection. By combining Ray's scalable orchestration with Gymnasium's standardized interface, Chrono-Gymnasium closes the usability gap between high-fidelity physics simulation and the distributed data pipelines required by modern RL and design optimization.


	\section{METHODS}

The Chrono-Gymnasium framework contains two main components. The first provides Gymnasium style interfaces as a specification for most Project Chrono problems. The second one provides an integration between Project Chrono and Ray to provide parallel computing capability.

\subsection{Chrono Simulation Scenario Modeling and Gymnasium Instrumentation}
\begin{figure}[htbp]
	\vspace{0.5em}
	\centering
	\includegraphics[width=\columnwidth]{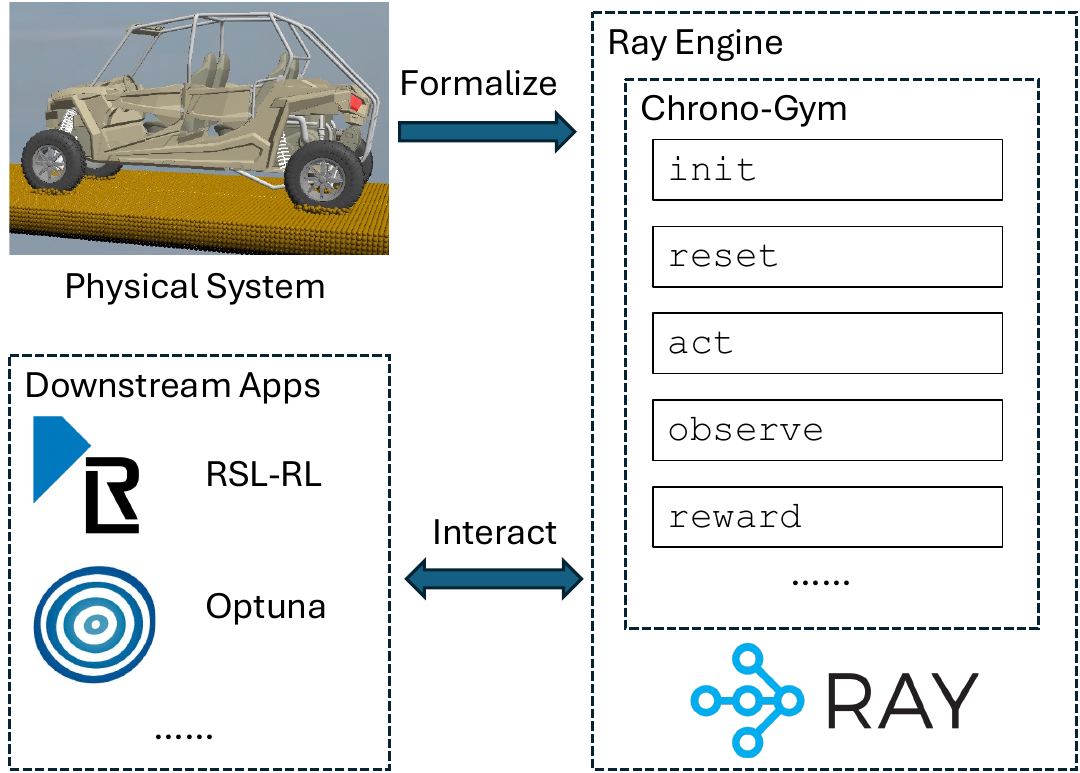}
	\caption{The architecture of Chrono-Gymnasium. Physical systems are described in a formalized way using Chrono-Gymnasium interfaces. Downstream applications like reinforcement learning and bayesian optimization can interact with the environment in a distributed manner through Chrono-Gymnasium Engine and Ray Engine }
	\label{fig:arch}
\end{figure}

Project Chrono simulation scenarios are usually written in the form of free-style C++ or Python programs, consisting of a series of Project Chrono API calls. We ``gymnasium instrument'' a Chrono virtual experiment as described below and shown in Figure \ref{fig:arch}. Note that the abstraction discussed below is agnostic to the specific RL solver, and this instrumentation is applicable to other classes of applications, e.g., Bayesian optimization, surrogate modeling, Bayesian calibration, which build on top of the Ray platform. We discuss RL since it is the most demanding framework in which Chrono Gymnasium is employed.

\begin{enumerate}
	\item \textbf{Simulation resources (initialization).}
	A Chrono simulation scenario typically comprises multiple objects (e.g., one or multiple agents, terrain, sensors, autonomy stacks) that must be created once and then reused. We model this as an initialization stage that constructs:
	\begin{itemize}
		\item a set of \emph{advanceable} objects (e.g. \texttt{ChSystem}, \texttt{ChFsiProblemSPH}) that must be stepped every simulation tick, and
		\item a set of additional objects/resources that are required for observation, control, or bookkeeping.
	\end{itemize}
	
	\item \textbf{Reset logic (initial conditions).}
	Resetting a Chrono simulation often means restoring initial positions/velocities, controller state, and any episode-specific randomization, while \emph{not} necessarily reloading heavy resources. We model reset as a separate callable that mutates the already-created objects according to a given seed and optional reset options.
	
	\item \textbf{Control interface (actions).}
	Many Chrono problems are controlled by setting actuator inputs (forces/torques), driver commands, or controller targets. We model this as an \emph{apply-action} function
	\[
	a_t \mapsto \text{(mutations of Chrono objects)},
	\]
	where $a_t \in \mathcal{A}$ is the action at decision step $t$.
	
	\item \textbf{Time integration (stepping / advancing).}
	Chrono objects do not all expose the same stepping method (e.g., a system may use \texttt{DoStepDynamics(dt)}, while other modules use \texttt{Advance(dt)} or \texttt{Update()}). To unify this, we model time integration as:
	\begin{itemize}
		\item a fixed simulation step size $\Delta t$,
		\item a decision-to-simulation ratio $k$ (steps per action), and
		\item a generic advance operator that is applied to every advanceable object.
	\end{itemize}
	Concretely, one environment decision step corresponds to $k$ Chrono integration ticks:
	\[
	\text{for } i=1\ldots k:\quad \text{advance\_all}(\text{advanceables}, \Delta t) \; .
	\]
	This separation is important in practice: controllers and learning algorithms may act at a lower frequency than the physics integrator for stability and performance.
	
	\item \textbf{State $\rightarrow$ observation mapping.}
	Chrono internally maintains a rich state (positions, velocities, contacts, sensor histories over a time grid). Most downstream solvers operate on a compact representation. We model this via an observation function
	\[
	o_t = g(\text{advanceables}, \text{objects}) \; ,
	\]
	producing $o_t \in \mathcal{O}$ (e.g., joint angles, angular velocities, end-effector pose, sensor readings).
	
	\item \textbf{Task objective (reward / cost).}
	Many problems define a performance measure such as tracking error, energy usage, stability, or progress. We model this as a scalar reward (or negative cost) function
	\[
	r_t = R(\text{advanceables}, \text{objects}, a_t, o_t) \; ,
	\]
	computed after stepping.
	
	\item \textbf{Episode logic (termination and truncation).}
	Chrono simulations typically require stopping criteria, e.g., falling over, leaving bounds, numerical instability, or reaching a goal. We split end-of-episode conditions into:
	\begin{itemize}
		\item \emph{termination}: the task is solved or has irrecoverably failed,
		\item \emph{truncation}: a time-limit or external cutoff ends the rollout.
	\end{itemize}
	This matches common evaluation practice and keeps the physics model independent from the training horizon.
	
\end{enumerate}

With this decomposition, a free-form Chrono script is reinterpreted as a collection of small, explicitly-scoped functions (init/reset/act/observe/advance/terminate), which provides a unified interface across diverse Chrono problems while preserving the flexibility of the underlying Chrono API.

\subsection{Parallelization}

We parallelize rollouts using Ray by encapsulating each Project Chrono environment in a Ray actor \cite{moritz2018ray} (one actor per simulation instance). Each actor owns and persists its local simulator state (e.g., dynamics system, assets, sensors, and buffers) and exposes a minimal remote interface -- primarily \texttt{reset()} and \texttt{step}$(a)$ -- so that a batch of environments can be advanced concurrently by issuing remote procedure calls.

A central driver process coordinates execution by broadcasting actions to all actors and collecting results with \texttt{ray.get}. This induces an explicit synchronization barrier: an environment step completes only after every worker has advanced and returned its outputs, keeping the rollout batch aligned in environment-step time. To reduce coordination and RPC overhead, we introduce a hyperparameter $n_s$. Rather than synchronizing at every physics integration timestep, each call to \texttt{step}$(a)$ advances the simulator for $n_s$ internal timesteps before returning. Increasing $n_s$ lowers communication frequency and improves throughput, while still producing a single aggregated transition per environment step.

\begin{figure}[htbp]
	\centering
	\includegraphics[width=\columnwidth]{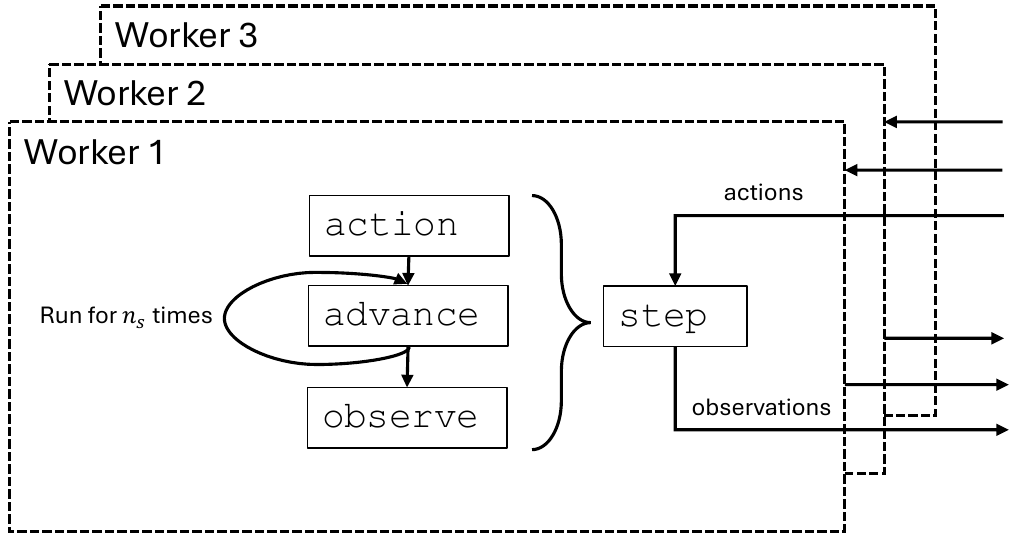}
	\caption{Actor (worker) step workflow for parallel Chrono environments. Each Ray worker keeps its simulator state resident; on every environment step it receives an action from the driver, applies it, advances the physics for  $n_s$ internal timesteps, then returns the resulting observations (and optional reward/termination/info). Sending actions and collecting outputs across multiple workers enables high-throughput. }
	\label{fig:workflow}
\end{figure}

To minimize data movement, large static assets remain resident within each actor. Only per-step outputs --observations (and, when applicable, rewards, termination flags, and auxiliary diagnostics)-- are serialized and returned to the driver as lightweight NumPy arrays and dictionaries. This design scales efficiently with the number of workers while preserving a straightforward, batch-oriented interface for reinforcement learning pipelines, see Figure \ref{fig:workflow}.

	\section{EXPERIMENT}

We present two distinct case studies. First, in the domain of robotics, we utilize Chrono-Gymnasium to train a reinforcement learning agent for Quadrupeds running on challenging terrains. Second, in engineering design, we apply the framework to the Bayesian Optimization of a planetary lander, where the high-fidelity contact dynamics of Chrono are essential for evaluating landing stability. For full implementation details and hyperparameter settings, please refer to our repository.

\subsection{Quadrupeds Locomotion -- RL}

\subsubsection{Task Description}
We consider a standard quadruped locomotion benchmark: \emph{commanded velocity-tracking} under episodic reinforcement learning. At each step, the policy receives the robot state and a desired planar base velocity command, and outputs joint-level actions to generate a stable walking gait. The objective is to track the commanded body velocity while maintaining balance and a nominal base posture throughout the episode.

An episode is deemed successful when the robot sustains stable locomotion for the full horizon without falling; episodes terminate early upon loss of balance (e.g., excessive roll/pitch) or other instability conditions. This task emphasizes closed-loop robustness: the policy must coordinate foothold timing and body stabilization to tolerate intermittent slippage, impacts, and state perturbations while continuing to track the commanded motion.

We choose velocity-tracking locomotion because it is a widely adopted reference task in modern quadruped RL benchmarks and is commonly provided as a baseline example in large-scale simulators (e.g., Isaac Lab~\cite{mittal2025isaac} and Genesis \cite{Genesis}). As a result, it serves as a meaningful point of comparison and a practical stress test for learning-based locomotion. Moreover, the task is contact-intensive and sensitive to modeling artifacts, making it a useful benchmark for evaluating whether a high-fidelity dynamics engine can support stable RL training without sacrificing locomotion performance.

\subsubsection{RL Setup}
\label{sec:rigid_rl}
We formulate locomotion as a continuous-control RL problem with an observation in $\mathbb{R}^{45}$ and an action in $\mathbb{R}^{12}$. The observation vector concatenates: 1) base angular velocity in the body frame ($\mathbb{R}^{3}$), 2) projected gravity in the body frame ($\mathbb{R}^{3}$), 3) commanded planar velocity ($\mathbb{R}^{3}$), 4) joint position offsets from a nominal posture ($\mathbb{R}^{12}$), 5) joint velocities ($\mathbb{R}^{12}$), and 6) the previous action ($\mathbb{R}^{12}$). Each group is scaled by a fixed factor to normalize magnitudes across physical quantities and improve learning stability. The action is a joint-space position target for all actuated joints ($\mathbb{R}^{12}$, three joints per leg). The policy outputs normalized actions that are linearly scaled and added to default joint angles before being applied to the simulator, following common quadruped RL practice where the network predicts joint targets around a nominal pose.

The per-step reward is a weighted sum of terms that encourage commanded-velocity tracking while regularizing stability and smoothness. The main component is an exponential tracking reward on the planar base velocity error. Additional terms penalize undesired yaw spin, vertical base velocity, deviation from a nominal base height, and large action-rate changes; a light joint regularization term keeps the posture close to the nominal configuration. 

We train the locomotion policy using Proximal Policy Optimization (PPO) \cite{schulman2017proximal} implemented in the RSL-RL library, together with the reward design described above. The policy follows a standard actor--critic formulation with MLP networks (hidden sizes [512, 256, 128], ELU activations) for both actor and critic. This PPO-based velocity-tracking setup is aligned with common practice in learning quadruped locomotion and provides a representative benchmark for evaluating RL performance~\cite{pmlr-v164-rudin22a, miki2022perceptive_loco, makoviychuk2021isaacgym, hoeller2024anymal_parkour}.

\subsubsection{Policy Performance}
\begin{figure*}[htbp]
	\centering
	\begin{subfigure}[b]{0.32\textwidth}
		\centering
		\includegraphics[width=\linewidth]{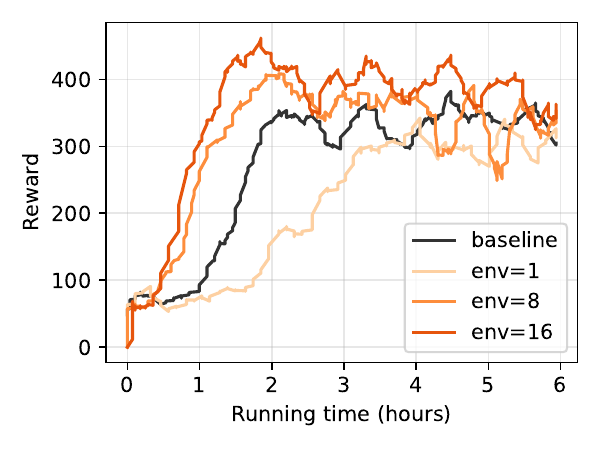}
		\caption{Reward - Wall Time}
		\label{fig:rl_reward}
	\end{subfigure}\hfill
	\begin{subfigure}[b]{0.32\textwidth}
		\centering
		\includegraphics[width=\linewidth]{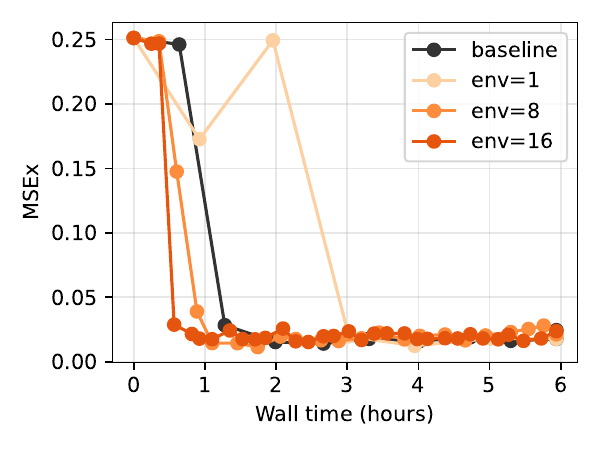}
		\caption{X MSE - Wall Time}
		\label{fig:rl_xmse}
	\end{subfigure}\hfill
	\begin{subfigure}[b]{0.32\textwidth}
		\centering
		\includegraphics[width=\linewidth]{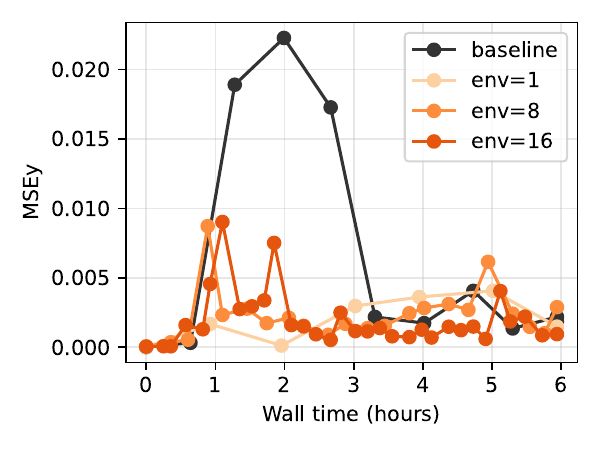}
		\caption{Y MSE - Wall Time}
		\label{fig:rl_ymse}
	\end{subfigure}
	
	\caption{Reinforcement-learning scalability with Chrono-Gymnasium. (a) PPO training reward versus wall-clock time; (b-c) episode-averaged velocity-tracking MSE in the forward (x) and lateral (y) directions. Results compare a serial Project Chrono baseline (no Ray) against Ray-based execution with $n_{env} \in \{1,8,16\}$ parallel environments. Increasing $n_{env}$ improves wall-clock learning efficiency and reduces tracking error; despite Ray's overhead at small scale, larger rollout parallelism yields around 2x overall speedup. }
	\label{fig:rl_exp}
\end{figure*}

As shown in Figure \ref{fig:rl_exp}, we evaluate the impact of parallel rollout scale on learning efficiency for the quadruped velocity-tracking task by varying the number of environments ($n_{\text{env}} \in \{1, 8, 16\}$). The figure reports (a) training reward as a function of \emph{wall-clock time}, and (b--c) the \emph{episode-averaged} mean-squared error (MSE) between the robot's base velocity and the target command in the $x$ and $y$ directions.

Across all metrics, increasing the number of parallel environments significantly improves wall-clock learning efficiency. In Figure \ref{fig:rl_reward}, the $n_{\text{env}}=16$ configuration converges the fastest and reaches the highest reward within the same time budget, while $n_{\text{env}}=8$ shows a similar trend with slightly slower convergence. In contrast, $n_{\text{env}}=1$ improves the slowest, requiring substantially more time to reach comparable returns. These results indicate that higher experience throughput translates directly into faster policy improvement in terms of wall-clock time.

The velocity-tracking errors in Figure \ref{fig:rl_exp}(b--c) are consistent with the reward trends. For the forward velocity ($x$ direction), $n_{\text{env}}=8$ and $16$ reduce the episode-averaged MSE rapidly and reach a low steady-state tracking error considerably earlier than $n_{\text{env}}=1$. A similar behavior is observed for lateral velocity tracking ($y$ direction), where increased parallelism again accelerates convergence to low MSE and suggests that the most highly parallel setting ($n_{\text{env}}=16$) yields the smallest lateral drift. Overall, the three subplots demonstrate that scaling the number of environments substantially reduces the wall-clock time required to learn accurate velocity tracking, yielding faster convergence and improved final performance on this locomotion benchmark.

\subsubsection{Transfer Learning to CRM Deformable Terrain}

To assess whether Chrono-Gymnasium can support learning beyond rigid-contact locomotion benchmarks, we perform a policy-adaptation experiment on deformable terrain using Chrono's Continuum Representation Method (CRM) \cite{weiGranularSPH2021,unjhawala2025physics}. A quadruped velocity-tracking policy is first trained on rigid terrain as described in previous section~\ref{sec:rigid_rl}, and the resulting PPO checkpoint is then used to initialize a second training stage in a CRM-based environment. The reward design, action, and observation interfaces are kept unchanged from the rigid-terrain task, so that the transfer experiment isolates the effect of terrain physics rather than changes in task specification or policy architecture. The CRM environment is implemented using \texttt{veh.CRMTerrain} and includes explicit SPH-based soil material and solver parameters, together with FSI registration of the robot feet and calf bodies. As a result, the robot interacts with terrain that exhibits compliance, sinkage, and deformation-dependent contact response. To maintain numerical stability, the CRM simulator is stepped with an internal timestep of $5 \times 10^{-4}$ s, and several such substeps are executed within each control interval.

Starting from the rigid-terrain checkpoint, PPO optimization is resumed with a conservative finetuning configuration: learning rate $10^{-5}$, clip parameter $0.02$, desired KL $0.001$, maximum gradient norm $0.5$, which supposed to avoid policy changed so drastically from rigid-trained baseline one. More broadly, the experiment demonstrates a practical simulator-fidelity curriculum in which locomotion structure is first learned efficiently on rigid terrain and then refined on a higher-fidelity deformable terrain model.

\begin{figure}[htbp]
	\centering
	\includegraphics[width=\columnwidth]{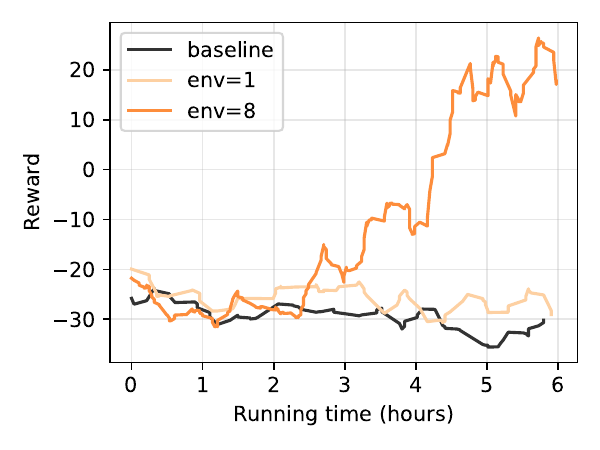}
	\caption{PPO training reward versus wall-clock time. Baseline refers to running a single experiment directly without Ray. }
	\label{fig:rl_exp_crm}
\end{figure}

As shown in Figure \ref{fig:rl_exp_crm}, the experiment results clearly show that Chrono-Gymnasium brings significantly speeds up the CRM finetuning process, making it possible to achieve good reward in a reasonable amount of time.

\subsection{Lander -- Bayesian Optimization}

\subsubsection{Task Description}

We consider the problem of optimizing the design parameters of a 
planetary lander to minimize impact loads during touchdown while 
maintaining structural stability.

The lander is modeled as a 25-body multibody system consisting of 
one chassis, six landing pads, and six articulated strut subsystems. 
Each strut subsystem is composed of three interconnected rigid bodies 
attached to the chassis through a combination of revolute, spherical, 
and prismatic joints.

The prismatic joint incorporates a custom nonlinear force model 
representing the honeycomb-based shock absorbers commonly used in 
planetary landers. This force model maps plastic deformation of the 
aluminum honeycomb structure to the corresponding axial reaction force, 
capturing the energy dissipation behavior during impact.

\begin{figure}[htbp]
	\centering
	\includegraphics[width=0.7\columnwidth]{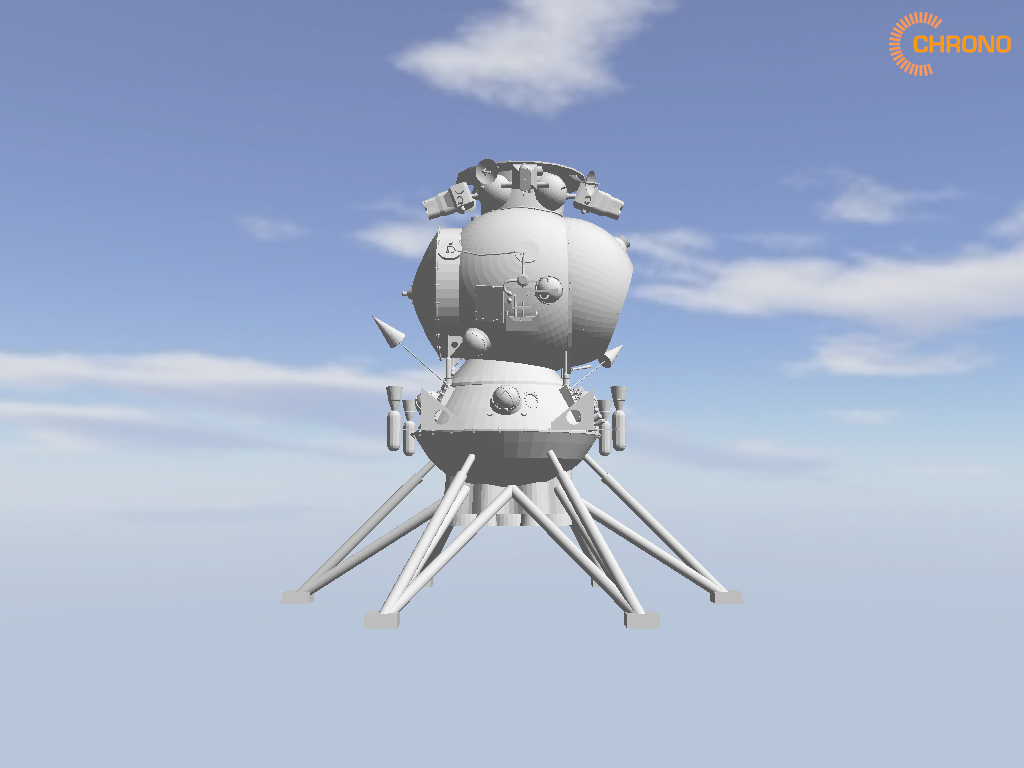}
	\caption{Rendered view of the Project Chrono planetary lander model used in the Bayesian Optimization case study. The 25-body multibody system comprises a rigid chassis and six landing legs with articulated strut assemblies, including a crushable honeycomb shock-absorber element to capture touchdown energy dissipation on deformable terrain. }
	\label{fig:lander}
\end{figure}

This modeling approach yields a high-fidelity digital twin suitable 
for simulation-driven design and optimization, and the experimental setup is shown in Figure \ref{fig:lander}. 

The touchdown phase is one of the most critical stages of a planetary 
mission. Excessive impact loads can damage instruments, compromise 
structural integrity, or cause vehicle instability on uneven or 
deformable terrain. 

Shock absorber stiffness therefore introduces a nonlinear trade-off: 
overly stiff designs increase peak deceleration, whereas overly 
compliant designs risk excessive stroke, bottoming-out, or reduced 
landing stability.

The design variables include the honeycomb yield force parameter $f_y$ 
and geometric suspension parameters such as strut inclination $\alpha_2$ and leg spread angles $\beta$ . We define the parameter vector as $\theta = \{ f_y,\; \beta,\; \alpha_2 \}$.

\subsubsection{Cost Function Overview}

Landing stability is quantified by the peak center-of-mass acceleration 
$a_{\max}$ experienced during impact.

To avoid the degenerate solution of infinitely stiff shock absorbers, 
we include an additional term that measures the fraction of the initial 
gravitational potential energy dissipated by the honeycomb structure.

The objective function is defined as
\begin{equation}
J(\theta)
=
\frac{a_{\max}}{a_{\mathrm{ref}}}
+
\left(
\alpha\left(
\frac{E_{\mathrm{abs}}}{E_{\mathrm{init}}}-1
\right)
\right)^2 .
\end{equation}

Here, $a_{\mathrm{ref}}$ is a reference deceleration associated with 
the maximum allowable deformation of the honeycomb structure.

$E_{\mathrm{abs}}$ denotes the energy dissipated by the shock absorbers, 
and $E_{\mathrm{init}}$ is the initial gravitational potential energy 
of the lander prior to impact. Lastly, $\alpha$ is a moderate weighting factor 
that penalizes deviations of the absorbed energy from the initial energy. 

We employ Bayesian Optimization (BO) to minimize $J(\theta)$ on both 
rigid and deformable terrain modeled using the CRM terramechanics framework.

\subsubsection{Prior Distributions}

We introduce empirical prior distributions as constraints to ensure that the optimization process generates meaningful designs.

\paragraph{Honeycomb Yield Force}

The parameter $f_y > 0$ controls the maximum reaction force of the 
crushable element. Impact deceleration scales approximately as
\[
a \sim \frac{f_y}{m},
\]
where $m$ is the lander mass.

The absorbed energy scales as $E_{\mathrm{abs}} \sim f_ys$ where $s$ is the crush stroke.

Because the system response depends multiplicatively on $f_y$, 
we adopt a log-uniform prior,
\[
f_y \sim \mathrm{LogUniform}(f_{\min}, f_{\max}),
\]
which assigns equal probability mass per order of magnitude 
and avoids bias toward excessively stiff shock absorbers.

\paragraph{Primary Strut Inclination}

The angle $\alpha_2$ determines the effective strut length and therefore 
strongly influences structural compliance and load distribution.

To avoid extreme configurations in which the strut becomes nearly 
horizontal or nearly vertical, we restrict $\alpha_2$ to a practical range 
and assume a bounded uniform prior with $0 < \alpha_{\min} < \alpha_{\max} < \tfrac{\pi}{2}$,
\[
\alpha_2 \sim \mathrm{Uniform}(\alpha_{\min}, \alpha_{\max}),
\]

This ensures physically meaningful and structurally feasible geometries.

\paragraph{Leg Spread Angle}

The parameter $\beta$ controls how far the supporting legs of the 
suspension system are spread relative to the strut centerline.

This geometric parameter influences global stability during ground contact, 
particularly on uneven terrain where asymmetric loading significantly 
affects the landing dynamics.

Since $\beta$ represents a purely geometric design choice, 
we again use a bounded uniform prior,
\[
\beta \sim \mathrm{Uniform}(\beta_{\min}, \beta_{\max}),
\]
defined over a practical range of angles.

Together, these priors define a realistic and physically admissible 
search space while remaining intentionally non-committal within 
reasonable engineering bounds. This enables Bayesian Optimization 
to explore suspension configurations efficiently without imposing 
strong structural bias.

\subsubsection{Optimization Implementation and Results}

We optimize the design vector $\theta = \{f_y, \beta, \alpha_2\}$ using Optuna~\cite{optuna_2019} with the Tree-structured Parzen Estimator (TPE) sampler~\cite{bergstra2011algorithms}. For each trial, the optimizer proposes a candidate lander configuration, the Chrono simulation evaluates the corresponding touchdown dynamics, and the resulting objective value $J(\theta)$ is returned to the optimizer. By executing these evaluations through Chrono-Gymnasium and Ray, multiple candidate designs can be assessed in parallel.

\begin{figure}[htbp]
	\centering
	\begin{subfigure}[b]{0.35\textwidth}
		\centering
		\includegraphics[width=\textwidth]{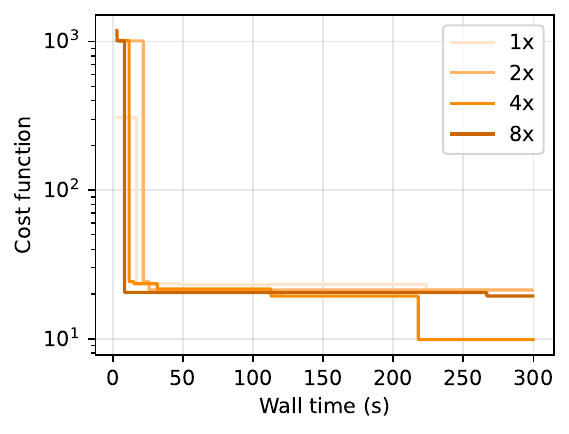}
		\caption{Rigid Terrain}
		\label{fig:bo_rigid}
	\end{subfigure}
	\hfill 
	\begin{subfigure}[b]{0.35\textwidth}
		\centering
		\includegraphics[width=\textwidth]{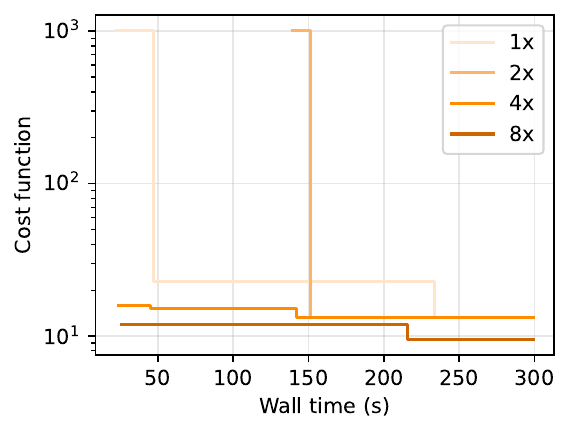}
		\caption{CRM Terrain}
		\label{fig:bo_crm}
	\end{subfigure}
	
\caption{Optimization scaling results on (a) rigid terrain and (b) CRM terrain. Curves show the best objective value found over wall-clock time for different parallel configurations. Increased parallelism reduces time-to-solution in both the CPU-bound and GPU-bound cases.}
	\label{fig:optimization_scaling}
\end{figure}

Figure~\ref{fig:optimization_scaling} shows the best objective value obtained as a function of wall-clock time for four parallel configurations on two terrain models: rigid terrain (Fig.~\ref{fig:bo_rigid}) and deformable CRM terrain\cite{unjhawala2025physics} (Fig.~\ref{fig:bo_crm}). In both studies, increasing the level of parallelism reduces the time required to identify low-cost designs. This trend is evident in the rigid-terrain experiment, which is primarily CPU-bound, and remains pronounced in the CRM-terrain experiment, which is primarily GPU-bound. In both cases, the more parallel configurations reach near-converged objective values substantially earlier than the lower-parallelism settings.

These results show that Chrono-Gymnasium, through its integration with Ray, can significantly accelerate simulation-driven design optimization across different computational regimes. 

	\section{CONCLUSIONS}
	
	This paper introduced Chrono-Gymnasium, an open-source distributed simulation framework that brings the high-fidelity multi-physics capabilities of Project Chrono into a Gymnasium-compatible interface and scales them across heterogeneous compute resources through Ray. By organizing Chrono-based problems around reusable components for initialization, reset, control, stepping, observation, reward, and termination, the framework reduces the engineering effort required to convert complex simulation models into environments suitable for reinforcement learning, Bayesian optimization, and other data-intensive workflows.

	The two case studies show that this abstraction is useful in practice across distinct application domains. In the quadruped locomotion benchmark, increasing the number of parallel environments improved wall-clock learning efficiency and reduced velocity-tracking error, demonstrating that high-fidelity Chrono models can support scalable policy training. In the planetary lander design study, distributed execution substantially reduced optimization time on both rigid and deformable terrain, highlighting the framework’s value for simulation-driven engineering design. Taken together, these results indicate that Chrono-Gymnasium helps close the gap between physically realistic simulation and the large-scale data generation required by modern learning and optimization pipelines.

	More broadly, Chrono-Gymnasium provides a practical path for combining simulation fidelity with distributed computation, without forcing users to build custom parallelization infrastructure for each new problem. Future work will focus on expanding support for additional Chrono application domains, improving execution efficiency for larger and more heterogeneous workloads, and broadening integration with downstream autonomy and design-optimization tools.
	
	\addtolength{\textheight}{-12cm}   
	


	
	
	\section*{ACKNOWLEDGMENT}
	
	This work was supported in part by the National Science Foundation under grants OAC-2519443 and CMMI-2153855.


	\bibliographystyle{IEEEtran}
	\bibliography{BibFiles/refsMBS,BibFiles/refsChronoSpecific,BibFiles/refsDEM,BibFiles/refsML-AI.bib, refs}

\end{document}